\documentclass[journal]{IEEEtran}
\ifCLASSINFOpdf
\else
\fi
\usepackage{amsmath}
\usepackage{booktabs}
\usepackage{xfrac}
\usepackage{graphicx}
\usepackage{tikz}
\usepackage{gensymb}
\usepackage{caption}
\usepackage{subcaption}
\usepackage{multirow}
\usepackage{diagbox}
\usepackage{makecell}
\usepackage{layouts}
\usepackage{ulem}
\usepackage{scalerel,amssymb}
\usepackage{mathtools}

\usepackage{nicefrac}

\begin{document}
%
\title{Towards intrinsic force sensing and control in parallel soft robots}
%
%
%

\author{Lukas Lindenroth, Danail Stoyanov, Kawal Rhode and Hongbin Liu
\thanks{This work has been conducted at the School of Engineering and Imaging Sciences, King's College London, UK. L. Lindenroth has been with the Department of Engineering, King's College London and is now with the Wellcome/EPSRC Centre for Interventional and Surgical Sciences (WEISS), University College London, 43-45 Foley St., Fitzrovia, London W1W 7EJ, UK}
\thanks{D. Stoyanov is with the Wellcome/EPSRC Centre for Interventional and Surgical Sciences (WEISS), University College London, 43-45 Foley St., Fitzrovia, London W1W 7EJ, UK}
\thanks{K. Rhode and H. Liu are with the School of Engineering and Imaging Sciences, King's College London, UK}
\thanks{Manuscript received -, 2021; revised -, 2021.}}

%
%

\markboth{}%
{Shell \MakeLowercase{\textit{et al.}}: Bare Demo of IEEEtran.cls for IEEE Journals}
%



\maketitle

\begin{abstract}
With soft robotics being increasingly employed in settings demanding high and controlled contact forces, recent research has demonstrated the use of soft robots to estimate or intrinsically sense forces without requiring external sensing mechanisms. Whilst this has mainly been shown in tendon-based continuum manipulators or deformable robots comprising of push-pull rod actuation, fluid drives still pose great challenges due to high actuation variability and nonlinear mechanical system responses. In this work we investigate the capabilities of a hydraulic, parallel soft robot to intrinsically sense and subsequently control contact forces. A comprehensive algorithm is derived for static, quasi-static and dynamic force sensing which relies on fluid volume and pressure information of the system. The algorithm is validated for a single degree-of-freedom soft fluidic actuator. Results indicate that axial forces acting on a single actuator can be estimated with an accuracy of 0.56 $\pm$ 0.66N within the validated range of 0 to 6N in a quasi-static configuration. The force sensing methodology is applied to force control in a single actuator as well as the coupled parallel robot. It can be seen that forces are accurately controllable for both systems, with the capability of controlling directional contact forces in case of the multi degree-of-freedom parallel soft robot.

\end{abstract}


%
\IEEEpeerreviewmaketitle

\section{Introduction}
%
%
%
%

\IEEEPARstart{S}{oft} robots have gained popularity following the paradigm shift in automation away from deterministic factory settings to collaborative, human-in-the-loop use-cases and medical procedures. Due to their compliance and softness when brought into contact with the environment soft robots are inherently safer given that high contact forces are mitigated through deformation around the point of contact. This has led to the development of a plethora of soft robotic systems with applications ranging from handling delicate objects \cite{Hughes2016} to minimally-invasive surgery \cite{Runciman2019}.

\begin{figure}[t!]
    \centering
    \includegraphics[width=\linewidth]{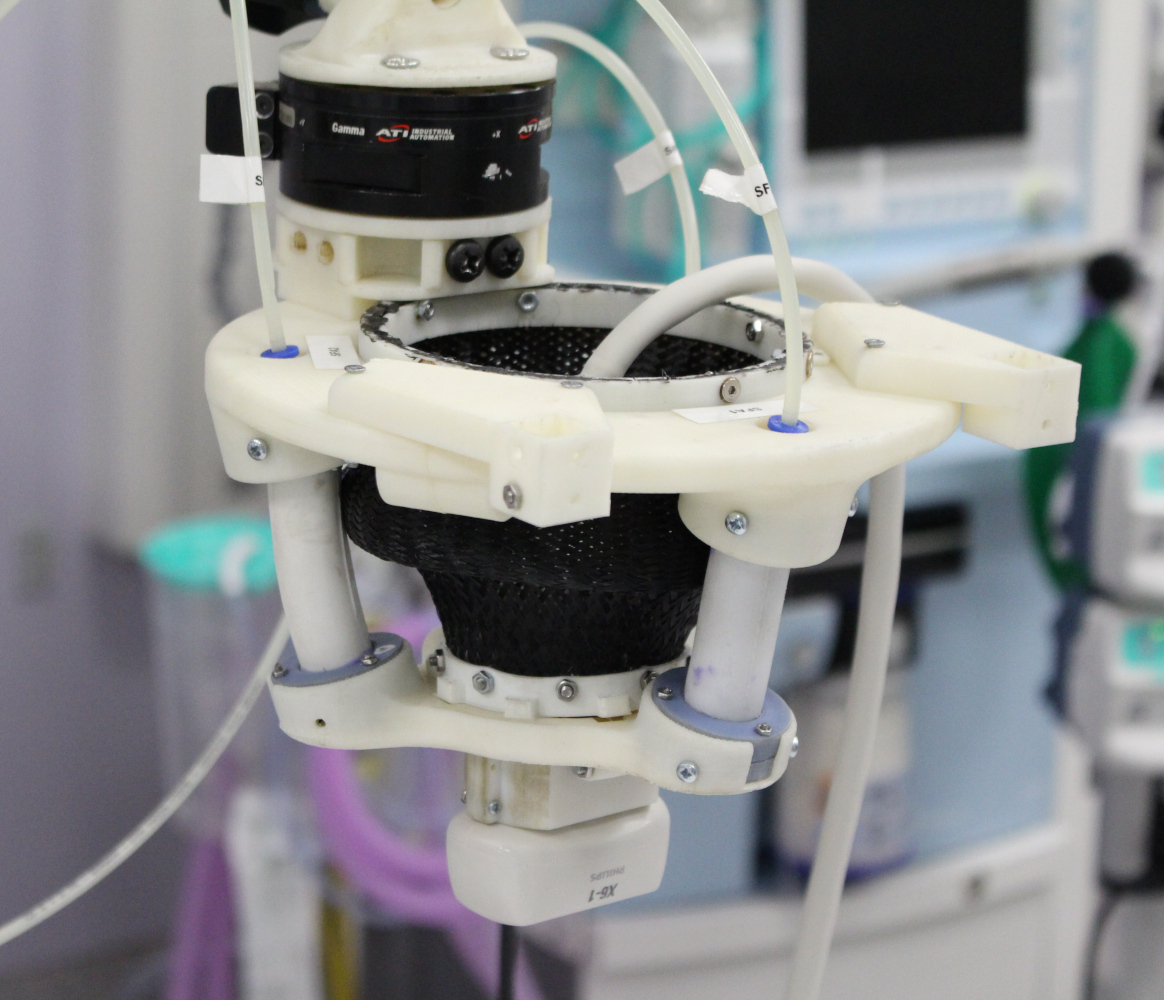}
    \caption{Soft robotic end-effector driven by extensible soft fluidic actuators equipped with an ultrasound probe}
    \label{fig:SEEoverview}
\end{figure}

Whilst there are a number of application scenarios in which high compliance of the robot is of paramount importance, significant deformability limits the usage of such robots for tasks requiring accurate positioning or precise loadbearing, as for example seen in catheter robotic systems \cite{Back2018} or soft robot grippers \cite{Galloway2016}. Thus, a soft robot has to be designed to exhibit the desired degree of compliance while conforming with task-specific requirements, which often leads to highly underactuated systems.

Different designs and methodologies have been derived to employ soft robotic systems in scenarios which require high loadbearing capabilities. Bishop-Moser proposes a toroidal fluidic actuator which offers significantly larger force ranges than a common McKibben actuator \cite{Bishop-Moser2019}. A high force 3d-printable soft bending actuator for use in a soft robotic hand exoskeleton has been proposed, yielding maximum forces 77.36N \cite{Yap2016}. Soft pneumatic actuators capable of induced high loads of $\approx$112N. Moreover, to increase the loading capabilities further the authors propose parallel arrangements of such actuators in packs. On a smaller scale, a film constraint-based a bendable soft micro-actuator has been proposed, capable of applying force up to 1.5N at a size of 16 mm × 40 mm × 850 µm \cite{Kosawa2019}. A novel design for a bellow-based soft robotic manipulator employing fluidic actuation capable of applying high loads has been presented in \cite{Zhang2020}. Such parallel topologies also find application in industrial systems, as for example shown by the Robotino® XT manipulator (Festo, Germany).

In our previous work we have shown that soft robots are particularly viable in medical applications, such as ultrasound imaging, and a parallel soft robotic system has been proposed which exhibit significant compliance in some DOFs while remaining stiff in others \cite{Lindenroth2017}. The system is designed around spring-reinforced soft fluidic actuators (SFAs) \cite{LindenrothTBME}. The soft robotic end-effector (SEE) system is capable of applying loads commonly found in ultrasound imaging without exhibiting substantial deformation in the contact force direction. Whilst the system has been proposed with the application of medical ultrasound in mind, such a soft parallel mechanism is highly adaptable and could be employed for a wide range of applications requiring the adaptability of a soft robot with the ability to control contact forces. An overview of the system equipped with an ultrasound probe is shown in Fig. \ref{fig:SEEoverview}.

To improve the load-bearing capabilities of a soft robot further, antagonistic stiffening and preloading mechanisms can be employed. Shiva et al. first proposed to antagonistically actuate tendons in a fluid driven soft robots to increase the systems stiffness by $\approx$100\% \cite{Shiva2016}. A backbone locking mechanism to increase the structural stiffness of a pneumatic soft robot with embedded elastic rods shows promising stiffness variation of $\approx$65\% \cite{Sun2020}. A bistable approach to high force and high speed soft robotic actuation has been proposed \cite{Tang2020}.

Such high force and high stiffness soft robotics systems can, in contrast to their highly-compliant counterparts, pose risks to their environments and, in case of medical robotic systems, patients. Thus, to keep system complexity minimal, integrated mechanisms for sensing and controlling applied contact forces have to be derived.

In current soft robotic systems force sensing is predominantly achieved by attaching dedicated force-torque sensors at the desired contact point on the robot's body or in close proximity to it. For this purpose, rigid, miniature force sensors have been developed and efficiently applied for example in medical procedures, minimally-invasive surgery and others. Different approaches have been developed based on physical quantities such as displacement and strain, resistance or pressure. An extensive review is given in \cite{Puangmali2008}. Tip force sensors are particularly employed in continuum systems \cite{Hoffmayer2015}. For the purpose of creating custom sensors, fibre optics show due to their flexible nature good applicability to measuring strains in sensing structures. Various fibre optics-based force sensors have been presented in \cite{Polygerinos2009, Kesner2011, Xiao2012, Polygerinos2013, Paridah2016}. A larger scale body contact sensor has been applied to a soft surgical manipulator in \cite{Xie2015}. It has been proposed to employ a rigid force sensor at the base of a soft robotic manipulator, thus maintaining a soft interaction \cite{Shiva2019}. The accuracy of estimating the force applied at a defined contact point along the robot's body is then highly dependent on the employed model as well as the feedback methodology for providing strain or deformation measures. Moreover, is the location of contact unknown or multiple forces act along the robot's body the approach fails.

Advances in material sciences have resulted in the development of flexible sensors which show the potential to be integrated in the elastomeric bodies of soft robots. Such sensors make use of e.g. piezoelectric polymers \cite{Qasaimeh2009, Seminara2013PiezoelectricSensors} or fluidic channels filled with liquid metal \cite{Chossat2013, Dickey2017, Cooper2017}. An overview on recent soft electronics sensors is given in \cite{Kramer2015}. 

While the above-mentioned concepts employ dedicated sensing devices attached to or embedded in the soft robotic structure, the robot could itself, due to its deformable and adaptive nature, be utilised to determine externally applied forces. This is commonly referred to as \textit{force estimation}. It has been shown that the information contained in the shape of the robot can be used to approximate the applied loading \cite{Khoshnam2015}. \cite{Rucker2011} present a probabilistic approach to force sensing in a continuum robot. A shape reconstruction algorithm with consecutive statics model is proposed in \cite{Yuan2017a}. A feature-based visual tracking algorithm combined with a finite element model for estimating external forces is presented in \cite{Zhang2018}. These approaches, however, require at least partial or in some cases full knowledge about the deformed shape of the manipulator, which is often limited either by the field of view or the update rate of the given sensing modalities. In \cite{Back2016} the external tip force acting on a steerable catheter robot has been investigated using electromagnetic marker tracking, indicating good applicability to clinical applications in which EM trackers are commonly used for catheter guidance.

Other investigations have shown that, besides the shape, the feedback in the actuator space of the robot could be utilised to determine externally applied forces, which is commonly referred to as \textit{intrinsic force sensing}. Extensive work has been conducted in intrinsically sensing the tip wrenches in multi-segment continuum robots \cite{Xu2008, Xu2010a}. In \cite{Haraguchi2011}, the authors propose a disturbance observer approach to determine external forces applied to a pneumatic manipulator, which is limited by the backbone material of the flexible joint required for the force sensing approach. \cite{Black2018} presents an actuator space-based force sensing approach for a parallel continuum robot. In our previous work we have shown that external forces applied to a hydraulic soft robotic manipulator can be related to internal hydraulic pressure variations in the actuating fluid given that an accurate model of the manipulator shape exists \cite{Lindenroth2017b}.



In this work we derive an intrinsic force controller which relies entirely on measurements of actuation pressures and induced fluid volumes. This is achieved by describing the intrinsic force sensing in the actuator space of the system and translating it into Cartesian space. Force sensing and control are derived for independent, single degree of freedom (DOF) soft fluidic actuators (SFAs) initially and combined for describing force control in a 3DOF parallel soft robotic end-effector (SEE), which is validated in clamping contacts. In summary, this work contributes the following.

\begin{itemize}
    \item Analysis of the internal pressure characteristics of an SFA under inflation and external loading, which enables accurate distinction between internal pressures caused by the SFA material strain and loads applied to the tip of the SFA.
    \item Force estimation in an SFA which enables the accurate intrinsic sensing of external forces under varying internal fluid volumes.
    \item A novel intrinsic force controller which relies solely on fluidic pressure and the known kinematic constraints applied to the system to sense and regulate external forces in a dynamic environment.
    \item Validation of the intrinsic force controller with a single SFA and the combined SEE with a stationary clamping contact under varying loads and inflation levels.
\end{itemize}

\section{Methodologies}

\begin{figure}
    \centering
    \includegraphics[width=\linewidth]{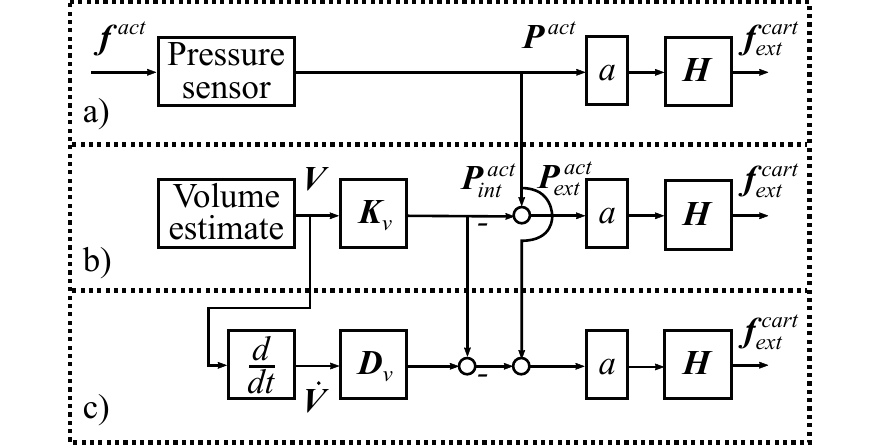}
    \caption{Overview of \textit{static} (a), \textit{quasi-static} (b) and \textit{dynamic} intrinsic force sensing algorithms (c)}
    \label{fig:Flowchart}
\end{figure}
\begin{figure}
    \centering
    \includegraphics{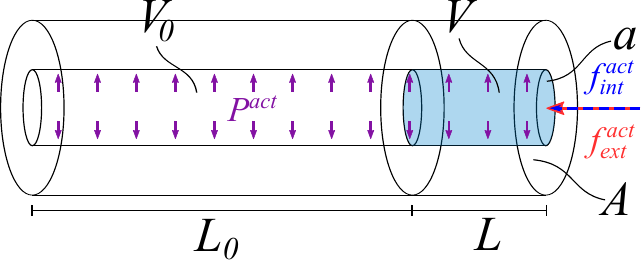}
    \caption{Free-body diagram of an SFA in deflated and inflated configurations under internal loading due to extension $f^{act}_{int}$ and external loading $f^{act}_{ext}$}
    \label{fig:drawing}
\end{figure}
Intrinsic force sensing in extensible soft robots is highly dependent on the robots' states and loading conditions. In the following, it is elaborated how static intrinsic sensing can be achieved when the robot configuration is ignored, which is demonstrated given the example of a single SFA. A lumped modelling approach is then described to consider configuration changes in the SFA. An approach to quasi-static and dynamic intrinsic force sensing is derived and force controllers are proposed for a single SFA and the SEE.

To simplify the derived control approach, force sensing and control are derived in the actuator space of the system. Forces are then related to the Cartesian space. Vectors containing forces and moments or kinematic information are denoted as $\ast^{act}$ and $\ast^{cart}$ when represented in actuator or Cartesian space respectively. An overview of the algorithms for \textit{static}, \textit{quasi-static} and \textit{dynamic} intrinsic force sensing is provided in Fig. \ref{fig:Flowchart}.

\subsection{Static force sensing}

In static force sensing the configuration-dependency of the internal fluid pressure is ignored. Here the robot is positioned in a given configuration and external forces are directly related to the pressure variation in the actuator space of the system. An external force applied along the direction of extension of the SFA induces a hydraulic pressure change of the form
\begin{equation}
    f^{act}_{ext} = P^{act}_{ext} \cdot a
    \label{eq:static}
\end{equation}
where $f^{act}_{ext}$ is the component of the external force acting along the direction of actuation of the SFA, $P^{act}_{ext}$ the resultant hydraulic pressure and $a$ the cross-sectional area of fluid channel.

This formulation can be expanded to a parallel assembly of multiple SFAs by determining the wrench transformation between a rigidly-connected frame and the individual SFAs, as it has been shown in our previous work for the previously-derived SEE \cite{LindenrothTBME}. Assuming a known, rigid transformation between actuators and a desired reference frame $\boldsymbol{H}$, the relationship can be written as

\begin{equation}
    \boldsymbol{f}^{cart}_{ext}= \boldsymbol{H} \cdot \boldsymbol{P}_{ext}^{act}\cdot a
\end{equation}
Where $\boldsymbol{f}^{cart}_{ext}\in[3\times1]$ is the external force applied in Cartesian space, $a$ the cross-sectional area of the actuator fluid chamber and $\boldsymbol{P}^{act}_{ext}\in[n\times1]$ the vector of external pressure applied to the $n$ individual actuators. $\boldsymbol{H}$ transforms the external force from actuator space into Cartesian space, such that
\begin{equation}
    \boldsymbol{H}=
    \begin{bmatrix}
    \nicefrac{\partial f^{cart}_x}{\partial f^{act}_1} & \nicefrac{\partial f^{cart}_x}{\partial f^{act}_2} & ... & \nicefrac{\partial f^{cart}_x}{\partial f^{act}_n}\\
    \nicefrac{\partial f^{cart}_y}{\partial f^{act}_1} & \nicefrac{\partial f^{cart}_y}{\partial f^{act}_2} & ... & \nicefrac{\partial f^{cart}_y}{\partial f^{act}_n}\\
    \nicefrac{\partial f^{cart}_z}{\partial f^{act}_1} & \nicefrac{\partial f^{cart}_z}{\partial f^{act}_2} & ... & \nicefrac{\partial f^{cart}_z}{\partial f^{act}_n}\\    
    \end{bmatrix}
\end{equation}

where $n$ is the number of actuators in the system. The formulation also holds for moments applied to the parallel structure, which will be ignored in this work given the limited number of SFAs employed in the SEE.

Whilst in an kinematically-constrained extensible actuator $a$ can be considered as constant, in reality the cross-sectional area of the fluid chamber changes under increasing fluid volume. Moreover, if the hydraulic fluid system contains entrapped air, the mechanical transmission between external force and hydraulic pressure change might be non-constant. Therefore, in practice, $a$ can be considered as the force transmission of the system and has to be approximated and experimentally determined.

\begin{figure}
    \centering
    \includegraphics[width=\linewidth]{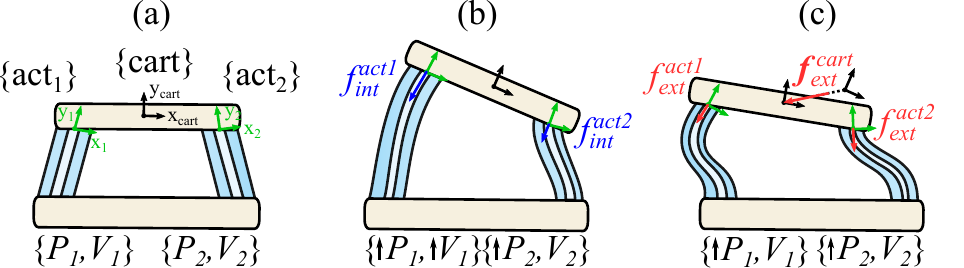}
    \caption{Configurations of the deflated SEE (a), under internal fluid volume variation, leading to internal reaction forces $f^{act_i}_{int}$(b) and under internal and external forces $\boldsymbol{f}_{ext}^{cart}$ (c).}    
    \label{fig:SeeConfOverview}
\end{figure}
\subsection{Quasi-static and dynamic force sensing}
Variation in the internal fluid volume $V$ leading to strain in the actuator induces stress which is reflected in form a hydraulic pressure $P_{int}$. Hence, changes in fluid volume within the actuators, and as such in the state of the system, need to be considered when estimating external forces.

The internal stress $\sigma$ exerted by an elastic body under an induced strain $\epsilon$ can be described
\begin{equation}
    \sigma = E\cdot \epsilon
\end{equation}
Where $E$ is the Young's Modulus of the elastic material. For a linear, isotropic elastic cylinder with an internal cavity of cross-sectional area $a$ as shown in Fig. \ref{fig:drawing}, a linearised local extension can be expressed as a function of the internal force $f^{act}_{int}$, such that
\begin{equation}
    f^{act}_{int} = \frac{EA}{L_0}L= \frac{EA}{V_0}V
\end{equation}
where $L_0$ and $V_0$ are the initial length and fluid volume of the actuator. The resultant internal pressure can be written as
\begin{equation}
    P^{act}_{int} = \frac{EA}{a^2 V_0} V = K_V V
\end{equation}

Where $K_V$ is the stiffness of the SFA.

If an external force is applied to the system along the direction of actuation then, under static equilibrium conditions, the following holds for the total hydraulic pressure $P^{act}$ in the system

\begin{equation}
    P^{act} = P^{act}_{int} + P^{act}_{ext}
\end{equation}
Where the total hydraulic pressure in the actuator is measurable by introducing a pressure transducer to the fluid system. To infer the external force applied axially to the SFA, one can employ the derived model such that
\begin{equation}
    P^{act}_{ext} = P^{act} - P^{act}_{int} = P^{act} - K_V V
\end{equation}
In case of the SEE, actuators are coupled in parallel. Thus, volume variation in a single SFA not only affects its internal pressure $P^{act}_{int}$ but the internal pressures and resulting forces of all coupled SFAs $\boldsymbol{f}^{act}_{int}$, as shown in Fig. \ref{fig:SeeConfOverview}. Expressing the stiffness for the combined mechanism therefore yields
\begin{equation}
    \boldsymbol{K}_V = 
    \begin{bmatrix}
    k_{11} & k_{12} & k_{13}\\
    k_{21} & k_{22} & k_{23}\\
    k_{31} & k_{32} & k_{33}
\end{bmatrix}
\end{equation}
Where $k_{ii}$ are the independent SFA stiffnesses and $k_{ij}$ the effects induced by volume variations of coupled SFAs. Whilst an analytical formulation can be found following the derivations shown in our previous work \cite{LindenrothTBME}, the elements of $\boldsymbol{K}_V$ not only depend on the elastic properties of the SFAs, but on the combined stiffness of actuators and fluidic drive system. The latter can be greatly affected by the chosen tubing and entrapped air amongst other factors. Therefore, in practice, $\boldsymbol{K}_V$ should be calibrated prior to sensing external forces.

The linear stiffness of the SFAs can be estimated by determining the pressure response of the system when inducing known fluid volumes $\boldsymbol{V}$. For a single DOF system, this can be achieved by inducing and retracting volume across the extension range of the actuator and linearising the pressure response. For the multi-DOF system, samples have to be acquired across the actuator space to consider configurations in which coupling forces occur between the actuators. The stiffness can then be estimated by solving the linear least-squares problem
\begin{equation}
    \underset{\boldsymbol{\hat{K}}_V}{\text{minimise}}\quad ||\boldsymbol{\hat{K}}_V \boldsymbol{V} - \boldsymbol{P}^{act}_{int}|| 
\end{equation}
Where $\boldsymbol{\hat{K}}_V$ is the approximated stiffness in the actuator space of the system,

This can then be related to the external force in Cartesian space by
\begin{equation}
    \boldsymbol{f}^{cart}_{ext} = \boldsymbol{H}(\boldsymbol{P}^{act} - \boldsymbol{K}_V \boldsymbol{V})\cdot a
\end{equation}

For fast SFA extensions, the damping of the system due to long fluidic lines becomes significant. Assuming linear damping with the damping coefficient ${D}_V$, the internal force of the actuator can then be expressed as 
\begin{equation}
    P^{act}_{int} = {K}_V {V} + {D}_V\dot{{V}}
\end{equation}
Where $\dot{{V}}$ expresses the actuation fluid flow rate.

For the combined SEE, the external force under consideration of damping caused by the fluid flow becomes
\begin{equation}
    \boldsymbol{f}^{cart}_{ext} = \boldsymbol{H}(\boldsymbol{P}^{act} - \boldsymbol{K}_V\boldsymbol{V} - \boldsymbol{D}_V\dot{\boldsymbol{V}})\cdot a
\end{equation}

\begin{figure}[t]
    \centering
    \includegraphics[width=\linewidth]{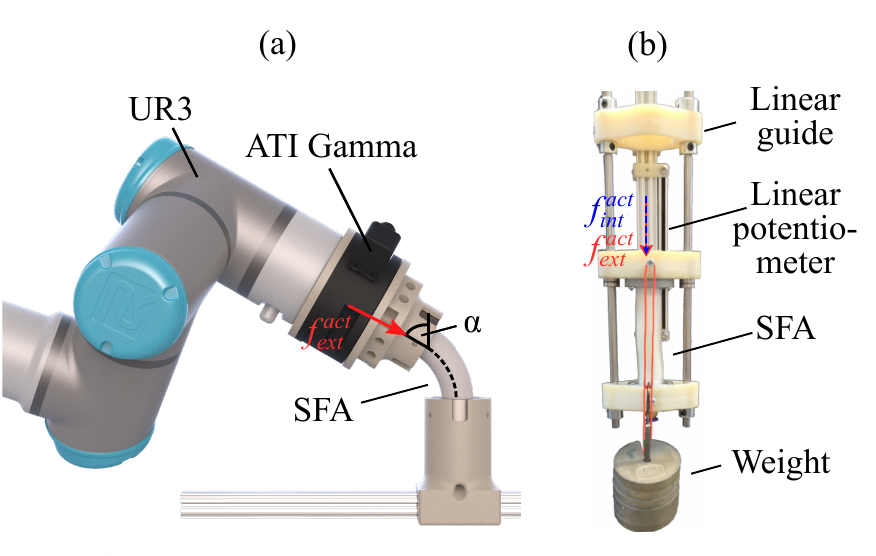}
    \caption{Experimental setups for SFA loading characterisation. The SFA tip is displaced by a robot manipulator (UR3) under a given bending angle $\alpha$ and the resulting force $f_{ext}^{act}$ measured by an inline force-torque sensor (ATI Gamma) (a). Weights are applied to the tip of the linearly-guided SFA to determine loading for different extensions (b).}
    \label{fig:Methods_SFA_Force}
\end{figure}

\subsection{Force control}

Based on the previously derived characteristics of the system, a linear force controller can be defined to drive the contact force to a desired force setpoint $\boldsymbol{f}^{cart}_d$ in Cartesian space.

For a single SFA, a general force PID controller can be defined as
\begin{equation}
    \dot{V}_d = G_{P} f^{act}_{e} + G_I \int_t f^{act}_{e} dt + G_{D} \dot{f}^{act}_{e}
\end{equation}
Where $G_{P}$, $G_{I}$ and $G_{D}$ are the proportional, integral and derivative gains respectively. $\dot{V}_d$ is the demanded SFA flow rate and $f^{act}_{e}$ denotes the force error in actuator space such that
\[
    f^{act}_{e} = f^{act}_d - f^{act}_{ext}
\]
With the demanded force $f^{act}_d$ along the central channel of the SFA. For the multi-DOF system with an arbitrary force direction in Cartesian space the force control law becomes

\begin{equation}
    \boldsymbol{\dot{V}}_d = \boldsymbol{G}_{P} \boldsymbol{f}_{e} + \boldsymbol{G}_I \int_t \boldsymbol{f}_{e} dt + \boldsymbol{G}_{D} \boldsymbol{\dot{f}}_{e}
\end{equation}

Where $\boldsymbol{G}_P$, $\boldsymbol{G}_I$ and $\boldsymbol{G}_D$ are diagonal gain matrices. The force error can then be defined in actuator space by

\begin{equation}
\boldsymbol{f}_e = \boldsymbol{f}^{act}_{d} - \boldsymbol{f}^{act}_{ext} = \boldsymbol{H}^{-1}\boldsymbol{f}^{cart}_{d} - \boldsymbol{P}^{act}_{ext}\cdot a
\end{equation}

with the demanded force in Cartesian space $\boldsymbol{f}^{cart}_{d} = [f_{d,x}, f_{d,y}, f_{d,z}]^T$.


\subsection{Experimental validation}

Experimental setups are created to validate the accuracy of the proposed intrinsic sensing and control approaches.

To determine the effect of the SFA state and shape on the force transmission of the system, it is characterised by applying axial loads and measuring the resulting hydraulic pressure variation. For this purpose the base of the SFA is fixed and its tip is slowly displaced by 2mm using a robot manipulator (UR3, Universal  Robots,  Odense,  Denmark). The resulting force applied to the SFA is measured with a force-torque sensor (Gamma,  ATI,  Apex,  USA). An overview of the setup is provided in Fig. \ref{fig:Methods_SFA_Force}a). The measurement is repeated for different levels of extension in 0.5ml volume and 10deg bending increments. Bending poses are obtained assuming a constant curvature arc given the length of the SFA and a desired tip bending angle. The force transmission is characterised for varying inflation volumes $V$ and bending angles $\alpha$.

\begin{figure}[t!]
    \centering
    \includegraphics[width=2.54in]{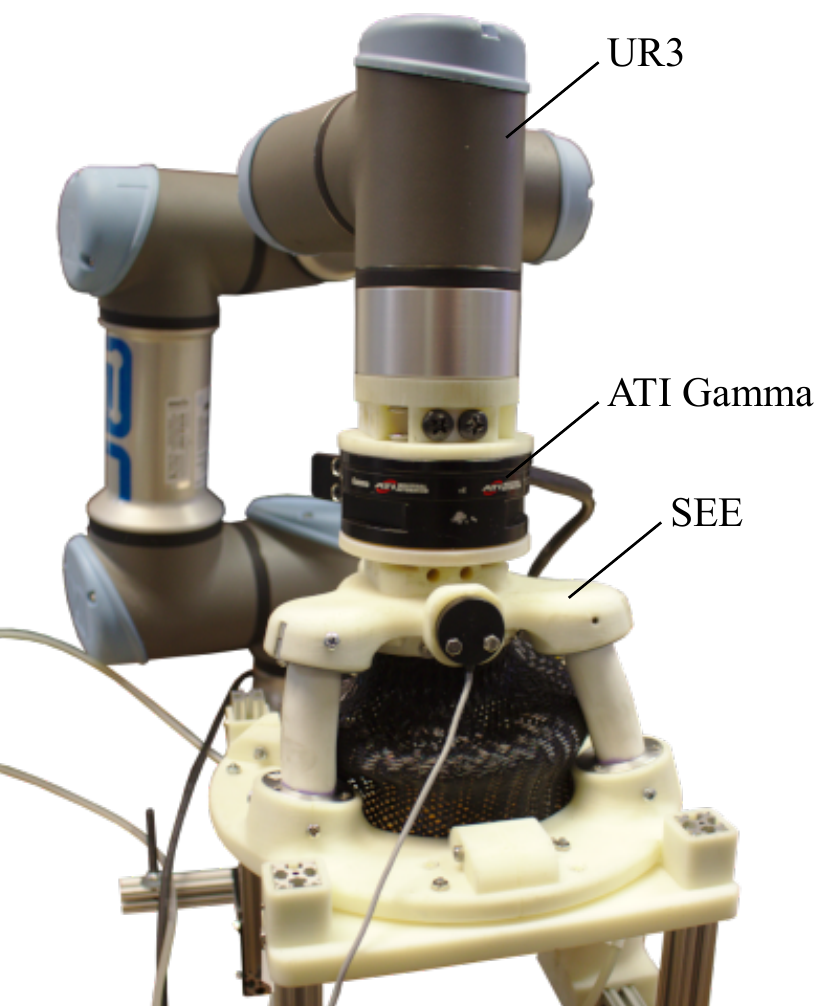}
    \caption{Experimental setup for force control evaluation with clamping contact provided by a robotic manipulator.}
    \label{fig:SREE_ForceControlOverview}
\end{figure}

To verify the pressure and extension behaviour of an SFA with a minimal influence of bending and shearing, a railed experimental setup is constructed. It comprises of two linear guides to which the SFA tip is connected. Two stainless steel rods are attached to an aluminium profile through two 3d-printed bracket elements. A sliding element is added which contains one linear bearing (LM6UU) per guide to reduce friction. The slider is connected to a linear potentiometer (100mm stroke length) which is fixed to the aluminium profile. The SFA is fixed to one of the bracket elements and attached to the slider. Upon inflation, the slider is moved and a proportional voltage is measurable across the potentiometer. The hydraulic pressure inside the fluid line is measured using a capacitive pressure transducer. A depiction of the experimental setup is presented in Fig. \ref{fig:Methods_SFA_Force}b).

The force controller for a single SFA as well as the integrated SEE are validated under a clamping contact constraint, which is achieved by affixing the base of the SFA and SEE and attaching the tip via a force-torque sensor to a robotic manipulator. Exemplary setups are shown for the single SFA in Fig. \ref{fig:Methods_SFA_Force}a) and for the SEE in Fig. \ref{fig:SREE_ForceControlOverview}. The individual SFAs are actuated through linear syringe pumps, as introduced in our previous work \cite{LindenrothTBME}. The controller is executed at a frequency of 250Hz and pressure data is filtered using an exponential moving average filter with a cutoff frequency of 100Hz.

\section{Results}

\subsection{Static force sensing}

\begin{figure}[t!]
    \centering
    \includegraphics[width=\linewidth]{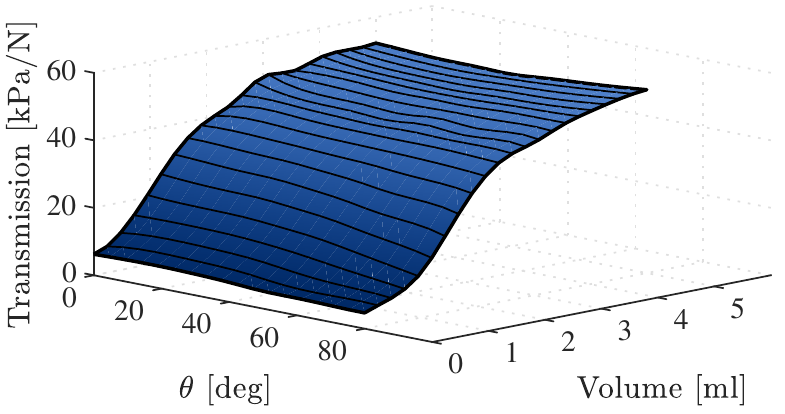}
    \caption{Force transmission of a single SFA with inflation and bending}
    \label{fig:SFA_Sensitivity}
\end{figure}

The force transmission as a function of the SFA inflation volume $V$ and bending angle $\alpha$ is presented in Fig. \ref{fig:SFA_Sensitivity}. It follows a highly nonlinear trend with the induced working fluid volume which approaches a constant value for volumes greater than 2.5ml. The effect of bending, however, is negligible compared to the sensitivity change due to extension. In the following, the SFA will be controlled in a region with minimal change to the transmission. To achieve this, only inflations $\geq2.5$ml are considered. For this range, the force transmission is 48.5$\pm3.25$kPa/N. The error introduced by linearizing the force transmission of the SFA is therefore 6.69\% of the maximum inflation pressure. The error introduced by bending the actuator is insignificant compared the change in inflation and is therefore ignored.

\begin{figure}
    \centering
    \includegraphics[width=\linewidth]{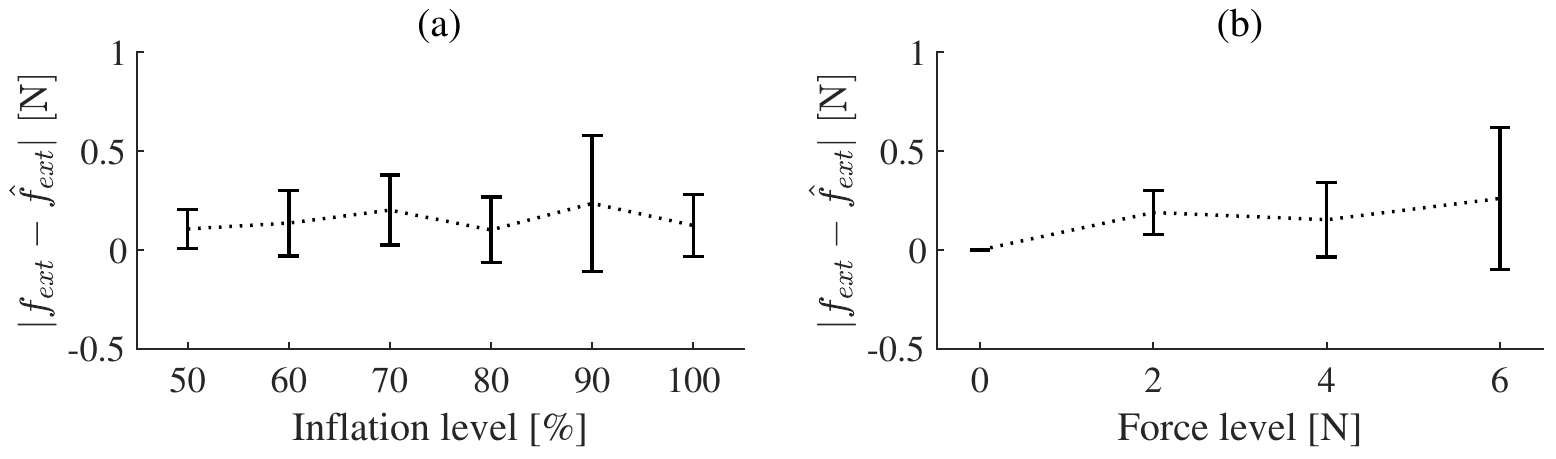}
        \caption{Force error in static intrinsic force sensing for different inflations (a) and loads (b) of a single SFA}
    \label{fig:SFA_StaticCalibration}
\end{figure}

The obtained force transmissibility is applied to verify the static and quasi-static force sensing capabilities of the SFA. In the static force sensing experiment, internal pressure $P^{act}_{int}$ is sensed once a desired inflation is reached and thus, pressure effects induced by the fluid volume are ignored. In quasi-static force sensing the internal pressure is estimated as a function of the induced fluid volume.

The SFA is attached to a linear guide and for different levels of extension, loads in form of weights are applied axially through a cable. Given the previously derived constant force transmission, the applied forces are estimated as indicated in Eq. \ref{eq:static}.

The SFA is inflated in a range from 50\% to 100\% in 10\% increments. For each inflation state, weights inducing loads of 2N, 4N and 6N are incrementally added to the tendon attached to the tip of the SFA before a new inflation volume is approached. The results are shown in Fig. \ref{fig:SFA_StaticCalibration}, where Fig. \ref{fig:SFA_StaticCalibration}a) shows the force error averaged over the applied loads for varying inflation and Fig. \ref{fig:SFA_StaticCalibration}b) the force error averaged over the different inflation levels for varying force levels. 

The results indicate that the inflation level has a less significant effect on the force sensing accuracy as the force level in static force sensing. Whilst the force error remains almost constant throughout different levels of inflation at 0.15$\pm$0.18N, the error due to the applied load increases from 0.18$\pm$0.11N for a load of 2N to 0.26$\pm$0.36N at 6N. It can be concluded that, without the contribution of uncertainty from randomised movements of the SFA, higher forces applied to the SFA lead to less accurate force sensing, whilst the inflation of the SFA does not contribute significantly to a change in error.

\begin{figure}[t!]
    \centering
    \includegraphics[width = 88mm]{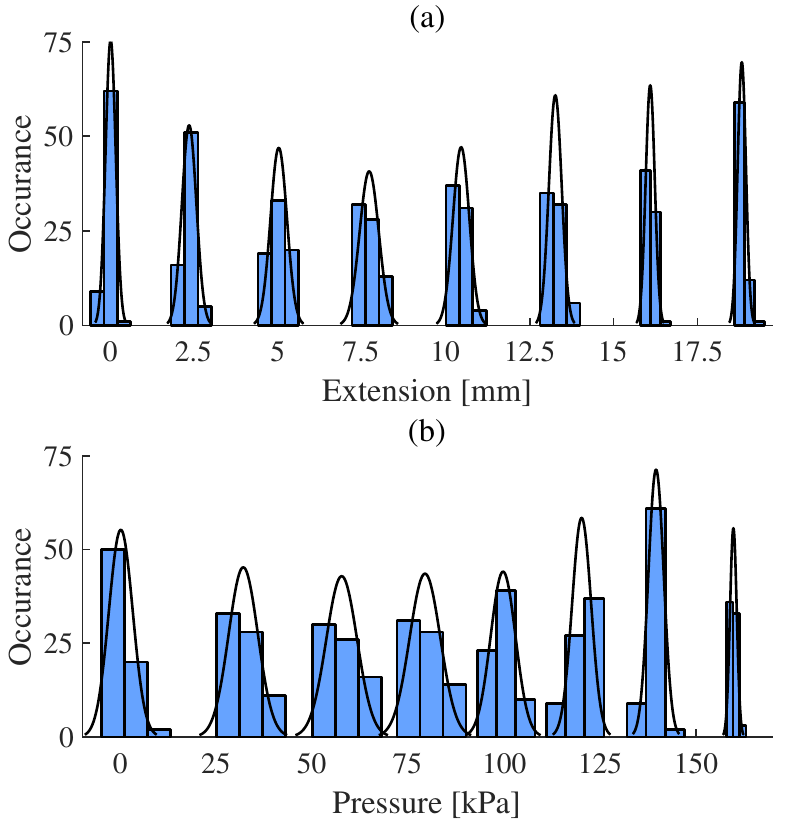}
    \caption[Variation in SFA extension and pressure for repeated volume variations]{Variation in SFA extension (a) and pressure (b) for repeated volume variations across 500 repetitions.}
    \label{fig:SFA_Repeatability}
\end{figure}

\subsection{Quasi-static force sensing}
\begin{figure}
    \centering
    \includegraphics[width = \linewidth]{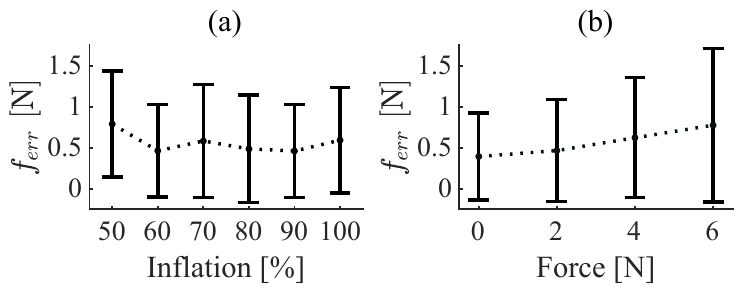}
    \caption[Force error in quasi-static force sensing for varying inflation and force levels]{Force error in quasi-static intrinsic force sensing for varying inflation (a) and force levels (b) of a single SFA}
    \label{fig:SFA_StaticValidation2}
\end{figure}

To demonstrate the effect caused by other unconsidered mechanical phenomena and to establish a maximum accuracy with which the internal pressure inside the SFA and thus the state of the SFA can be estimated as a function of the induced fluid volume, a repeatability experiment is conducted. To this end, a set amounts of working fluid are induced into the SFA and the extension and pressure responses are recorded in steady state. To ensure consistency of the results, this is repeated 500 times. The order of the demanded volumes is randomised to show the effect of history-dependent phenomena such as hysteresis.

Results of the experimental validation are shown in Fig. \ref{fig:SFA_Repeatability}. The standard deviations for each SFA state are presented in Table \ref{tab:SFA_Repeatability}. The accuracy in both the SFA length and pressure vary substantially with the induced extension. While the positioning uncertainty reaches local minimum of 0.15mm and 0.12mm at the 0\% and 100\% of volume, a maximum standard deviation 0.29mm is observed at 42.9\% extension. A similar trend is observable in the pressure sensing. The pressure sensing uncertainty reaches its minimum at 100\% extension and a maximum of 4.20kPa at 28.57\% extension. The mean SFA state uncertainty is 0.2mm in position and 2.95kPa in pressure. Taking into account the force transmissibility of the system, this relates to an uncertainty in force of 0.06N.
\begin{figure}[t!]
    \centering
    \includegraphics[width = \linewidth]{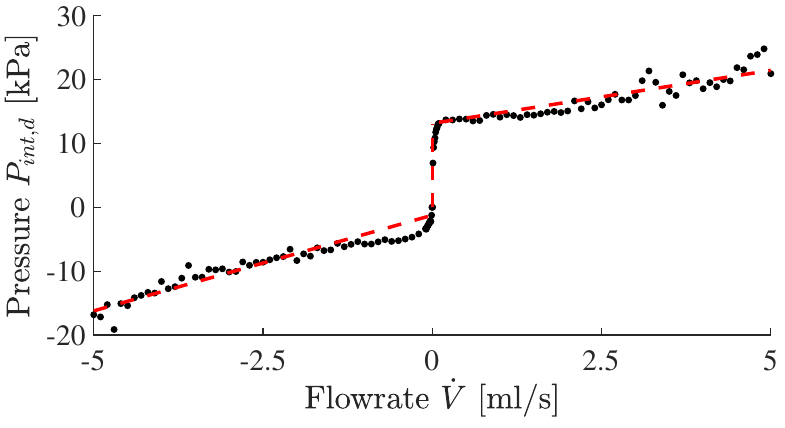}
    \caption{Damping pressure for different flow rates $\dot{V}$ of a single SFA}
    \label{fig:SFA_Damping}
\end{figure}

\begin{figure*}[t!]
    \centering
    \includegraphics[width = \textwidth]{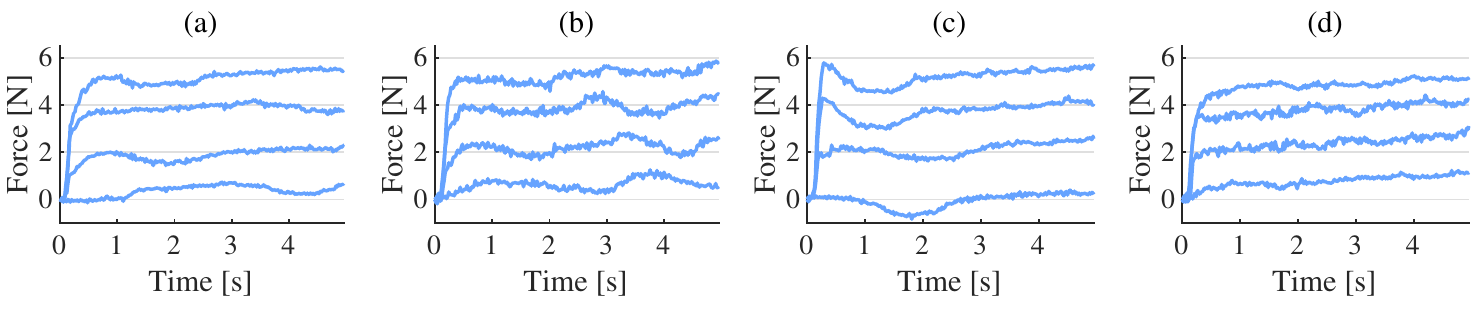}
    \caption{Results for SFA step responses at 50\% (a), 60\% (b), 70\% (c) and 80\% (d) initial inflation showing the measured forces for varying force demands of 0N, 2N, 4N and 6N.}
    \label{fig:SFA_ForceControl}
\end{figure*}




\begin{table}[h!]
\centering
\caption{State estimation accuracy}
\label{tab:SFA_Repeatability}
\begin{tabular}{ccc}
\toprule
Volume $[\%]$ & $\sigma(x) [\text{mm}]$ & $\sigma(P) [\text{kPa}]$ \\ \midrule
0       &   0.15    &   3.12 \\
14.29   &   0.22    &   3.81 \\
28.57   &   0.25    &   4.02 \\
42.90    &   0.29    &   4.01 \\
57.14   &   0.24    &   3.30  \\
71.43   &   0.19    &   2.49 \\
85.71   &   0.14    &   2.01 \\
100.0     &   0.12    &   0.88 \\
\midrule
$\mu$   &   0.20     &   2.95 \\
\bottomrule
\end{tabular}
\end{table}
An experimental validation of the quasi-static force sensing capabilities of the SFA is conducted. A randomised sequence of 100 poses in the range from 50\% to 100\% extension with increments of 10\% is generated. A random load of 0N, 2N, 4N or 6N is assigned to each pose and attached to the SFA once the pose target is reached and dynamic pressure variations have settled. The resulting data are summarised for varying inflation level in Fig. \ref{fig:SFA_StaticValidation2}a) and for varying force level in Fig. \ref{fig:SFA_StaticValidation2}b).


In comparison to the results of the static force sensing experiment, the mean error across all configurations is increased from 0.15$\pm$0.18N to 0.56$\pm$0.66N. Similar to the previous results, it can be seen that the accuracy of the estimation decreases with the applied load, from 0.39$\pm$0.53N at 0N load to 0.77$\pm$0.94N at 6N. While the accuracy decreases with the applied load, it is only minimally affected by the inflation level. The external force applied to the SFA can be estimated with an average accuracy of 0.56N$\pm$0.66N across all configurations.

To be able to compensate for pressure variation due to high fluid flow rates, an experiment is conducted to determine the pressure drop associated with a given flow rate. The SFA is attached to the linear guide and inflated with a known flow rate. Upon reaching its maximum inflation, it is deflated with the same flow rate. Flow rates are varied between 0ml/s and 5ml/s, and the resultant damping coefficient $D_V$ computed such that

\begin{equation}
\label{eq:SFA_DampingCoefficient}
    D_V = \frac{P-K_V V}{\dot{V}}
\end{equation}

50\% of the beginning of each trajectory are ignored due to pressure fluctuations induced by the rapid acceleration of the system. The results of the evaluation are presented in Fig. \ref{fig:SFA_Damping}, where each sample corresponds to the computed damping pressure $P_{int,d} = D_V \dot{V}$ for a given constant flow rate of the SFA across one inflation and deflation cycle. The pressure offset caused by the damping of the system is highly discontinuous and nonlinear. It can be described by a discontinuous function of the form

\begin{equation}
D_V = 
    \begin{dcases}
        3.00\frac{kPa}{ml/s} - 16.82kPa &, \dot{V} \leq 0 \\
        13.1kPa &, 0ml/s < \dot{V}\leq 0.1ml/s \\
        1.66\frac{kPa}{ml/s} + 13.10kPa&, 0.1ml/s < \dot{V}\\
    \end{dcases}
\end{equation}

To simplify the controller and avoid discontinuities, the combined damping as a function of the flow rate is linearised across the investigated flowrate ranges. For the given SFA, the linearised damping is determined as 4.46$\frac{kPa}{ml/s}$.



\subsection{SFA force control}
\label{SFA_ForceControl}
To verify the performance of the force controller, step responses of the system are recorded for steps with demand forces of 0N, 2N, 4N and 6N of axial force. For that purpose the base and tip are fixed in space before the step is demanded. Due to the possible variability of the force controller performance with the extension length of the SFA, data is recorded for initial inflation volumes $V_{init}$ of 50\%, 60\%, 70\% and 80\%. For the single SFA, a \textit{PD}-controller with $G_P = 1.97 \frac{\text{ml}}{\text{s}\cdot \text{N}} $ and $G_D = 0.2 \frac{\text{ml}}{\text{s}^2\cdot \text{N}}$. Upon reaching the demanded force, the SFA is deflated, inflated to $V_{init}$ and the next target force is approached.

Time series of the step responses are shown in Fig. \ref{fig:SFA_ForceControl} and the resulting steady state errors are summarised in Table \ref{tab:SFA_ForceControl} below.


\begin{table}[h!]
  \centering
  \caption[Steady state error of the SFA force controller]{Steady state error of the SFA force controller for varying inflation levels and target forces.}
    \begin{tabular}{cccccc}\toprule
    \diaghead{$V_{init}$ $f^{act}_d$}{$V_{init}$}{$f^{act}_d$}& 0N & 2N & 4N & 6N & $\mu$\\\midrule
    50\% & 0.48N & 0.20N & 0.13N & 0.74N & 0.39N\\
    60\% & 0.67N & 0.36N & 0.26N & 0.59N & 0.47N\\
    70\% & 0.27N & 0.35N & 0.18N & 0.61N & 0.35N\\
    80\% & 0.93N & 0.55N & 0.18N & 1.02N & 0.67N\\\midrule
    $\mu$& 0.59N & 0.36N & 0.19N & 0.74N & 0.47N\\\bottomrule
    \end{tabular}%
  \label{tab:SFA_ForceControl}%
\end{table}%

\begin{figure*}[t!]
    \centering
    \includegraphics[width = \textwidth]{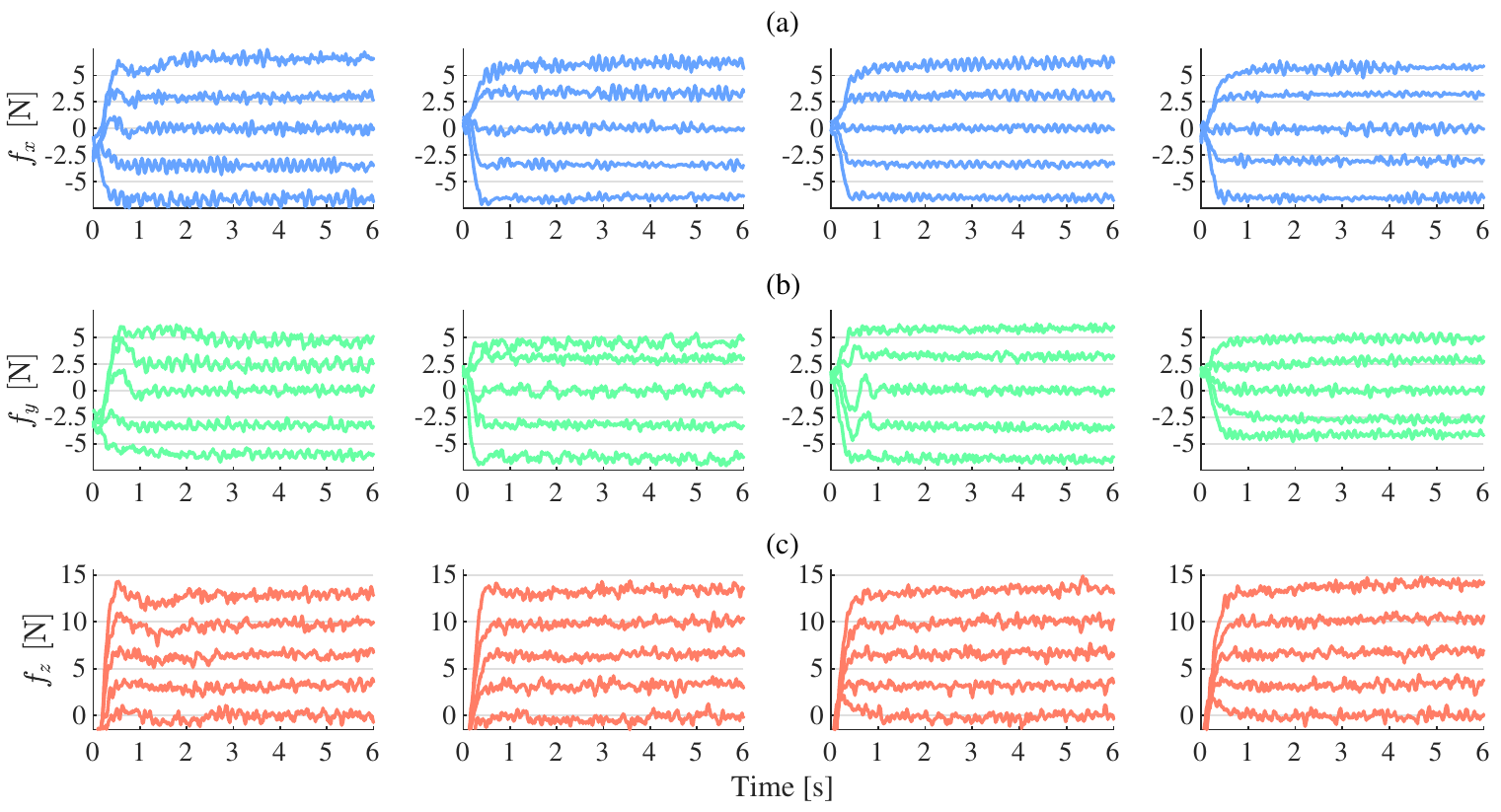}
    \caption{Results for SEE step responses at 50\%, 60\%, 70\% and 80\% initial inflation indicating the measured forces for varying demanded forces of -5N, -2.5N, 0N, 2.5 and 5N in $x$- (a) and $y$-axis (b) as well as 0N, 5N, 10N and 15N in $z$-axis (c).}
    \label{fig:SEEForceControl}
\end{figure*}





The controller achieves the desired forces well with a mean absolute steady state error of 0.47$\pm$0.24N across all configurations and force levels. This is in range with the previously validated quasi-static force sensing accuracy. The precision of the controller remains consistent over all tested configurations. Best results are achieved for demand forces of 4N and inflations of 70\% and 80\%. For most test conditions a time-dependent change in the steady state error is observable. This is particularly the case for large forces and large inflations. The contact force increases over time whereas the estimated force remains constant. This could be caused by the stress relaxation of the elastic material. Stress relaxation leads to a slow decrease in the internal stress, which in return results in a decreasing estimate of the external force. The controller compensates the falsely measured force difference, resulting in an increasingly higher external force.

\subsection{SEE force control}


Prior to evaluating the force controller performance, a calibration is undertaken to determine $\boldsymbol{K}_V$. A typical estimate $\boldsymbol{\hat{K}}_V$ is shown below.
\begin{equation}
    \boldsymbol{\hat{K}}_V = 
    \begin{bmatrix}
    43.31 & 1.94 & 1.48\\
    0.94 & 48.64 & 1.39\\
    0.36 & 1.18 & 44.18
\end{bmatrix} \textrm{ with } [\boldsymbol{\hat{K}}_V] = \frac{kPa}{ml}
\end{equation}
The force controller is implemented as a \textit{PI} controller with gains $G_P = 1.97 \frac{\text{ml}}{\text{s}\cdot \text{N}} $ and $G_I = 0.02 \frac{\text{ml}\cdot \text{s}}{\text{s}\cdot \text{N}}$
The step responses of the combined SEE for different inflation levels and demand forces are evaluated. Steps are given for forces in X-, Y- and Z-direction. The responses are shown in Fig. \ref{fig:SEEForceControl} and steady state errors are summarised in Tables III, IV and V for measured forces $f_x$, $f_y$ and $f_z$.

The measured force responses show good controllability of the tip force. Larger steady state force deviations occur only for demanded forces with greater magnitudes of $|f_d|\geq2.5$N for X- and Y- and $f_d\geq10$N for Z-direction. Across all directions and demanded forces it can be seen that force error decreases significantly with the initial inflation of the SEE. Whilst for 50\% inflation the mean steady state error is 0.98N, it decreases by 29\% to 0.7N at 80\% inflation. The error scales with the magnitude of the force. Similar diverging behaviour as seen in the SFA step responses in steady state is mainly observable for large demand forces. Its impact on the controller is significantly lower compared to the controller performance for a single SFA. 

\begin{table}[h!]
  \centering
  \caption{SEE force control steady state error for different inflation levels and target forces in X-direction}
  \resizebox{\linewidth}{!}{  
    \begin{tabular}{ccccccc}\toprule
     \diaghead{$V_{init}$ $f^{cart}_{d,x}$}{$V_{init}$}{$f^{cart}_{d,x}$} & -5N & -2.5N & 0N & 2.5N & 5N & $\mu$\ \\\midrule
    50\% & 1.66$\pm$0.63N & 0.98$\pm$0.56N & 0.47$\pm$0.55N & 0.55$\pm$0.48N & 1.56$\pm$0.56N & 1.04$\pm$0.55\\
    60\% & 1.52$\pm$0.37N & 0.94$\pm$0.35N & 0.36$\pm$0.44N & 0.84$\pm$0.55N & 1.12$\pm$0.62N & 0.96$\pm$0.42\\
    70\% & 1.53$\pm$0.36N & 0.89$\pm$0.33N & 0.31$\pm$0.35N & 0.65$\pm$0.48N & 1.11$\pm$0.55N & 0.90$\pm$0.46\\
    80\% & 1.56$\pm$0.36N & 0.57$\pm$0.43N & 0.39$\pm$0.48N & 0.72$\pm$0.26N & 0.73$\pm$0.43N & 0.79$\pm$0.45\\\bottomrule
    \end{tabular}}
    
  \label{tab:SEEForceFx}%
\end{table}%

\begin{table}[h!]
  \centering
    \caption{SEE force control steady state error for different inflation levels and target forces in Y-direction}  
  \resizebox{\linewidth}{!}{
    \begin{tabular}{ccccccc}\toprule
    \diaghead{$V_{init}$ $f^{cart}_{d,y}$}{$V_{init}$}{$f^{cart}_{d,y}$} & -5N & -2.5N & 0N & 2.5N & 5N & $\mu$\\\midrule
    50\% & 1.05$\pm$0.46N & 0.80$\pm$0.50N & 0.43$\pm$0.53N & 0.57$\pm$0.65N & 0.57$\pm$0.68N & 0.68$\pm$0.24\\
    60\% & 1.39$\pm$0.49N & 0.78$\pm$0.44N & 0.41$\pm$0.52N & 0.56$\pm$0.45N & 0.60$\pm$0.51N & 0.75$\pm$0.38\\
    70\% & 1.45$\pm$0.44N & 0.96$\pm$0.42N & 0.35$\pm$0.43N & 0.73$\pm$0.44N & 0.82$\pm$0.40N & 0.86$\pm$0.40\\
    80\% & 0.87$\pm$0.42N & 0.40$\pm$0.44N & 0.37$\pm$0.45N & 0.43$\pm$0.45N & 0.36$\pm$0.42N & 0.49$\pm$0.22\\\bottomrule
    \end{tabular}}
  \label{tab:SEEForceFy}
\end{table}%

\begin{table}[h!]
  \centering
  \caption{SEE force control steady state error for different inflation levels and target forces in Z-direction}
  \resizebox{\linewidth}{!}{  
    \begin{tabular}{ccccccc}\toprule
    \diaghead{$V_{init}$ $f^{cart}_{d,z}$}{$V_{init}$}{$f^{cart}_{d,z}$} & 0N & 3.75N & 7.5N & 11.25N & 15N & $\mu$\\\midrule
    50\% & 0.59$\pm$0.76N & 0.72$\pm$0.68N & 1.10$\pm$0.62N & 1.54$\pm$0.67N & 2.10$\pm$0.67N & 1.21$\pm$0.62N\\
    60\% & 0.56$\pm$0.67N & 0.66$\pm$0.67N & 1.02$\pm$0.61N & 1.32$\pm$0.68N & 1.64$\pm$0.67N & 1.04$\pm$0.45N\\
    70\% & 0.48$\pm$0.63N & 0.65$\pm$0.60N & 0.91$\pm$0.62N & 1.24$\pm$0.66N & 1.54$\pm$0.65N & 0.96$\pm$0.43N\\
    80\% & 0.49$\pm$0.65N & 0.65$\pm$0.65N & 0.82$\pm$0.62N & 1.00$\pm$0.62N & 1.11$\pm$0.66N & 0.82$\pm$0.25N\\\bottomrule
    \end{tabular}}
  \label{tab:SEEForceFz}%
\end{table}%

\section{Discussion}

This work establishes a practical approach to intrinsic force sensing and subsequent force control in fluid-driven soft robotic systems.

From the mechanical characterisation of the system it can be seen that a wide range of physical phenomena have to be taken into account to produce a viable model of the system across its entire workspace. This includes a non-constant, nonlinear force transmission, history-dependent phenomena which induce pressure and consequently force uncertainty and dynamic effects such as damping. In this work, we determine a range of SFA inflation which allows for simplification and linearisation of these phenomena. It can be seen that these metrics become more consistent upon inducing initial strain in the material.

We have confirmed that it is possible to sense and control external contact forces when the induced working fluid volume and pressure are known. Repeated online calibration of the system is necessary to capture adverse effects such as air bubble formation within the fluid-lines after the initial characterisation of the system. Consistent force sensing is then possible across the previously defined range.

It has been shown that intrinsic force sensing in a quasi-static configuration is possible with an average accuracy of 0.56$\pm$0.66N. As indicated by the experimental results, various factors contribute to the deterioration of the force sensing accuracy. Hysteresis for example causes history-dependent deviations from the linearised volume-pressure relationship. As the hysteretic behaviour of the system is not modelled, this can lead to errors in the prediction of the internal pressure and thus affect both the force sensing and controller performances.

A time-dependent deterioration of the steady state response of both the SFA and SEE force controllers is observable. One factor which greatly contributes to a time-dependent pressure change in an SFA is the relaxation of the utilised elastic silicone rubber material. In some of the SFA and SEE step responses a slow divergence from the demanded force value is noticeable. The relaxation of the material leads to a slow decrease of measured internal pressure in the system. This, in return, leads to an increasing difference between estimated and true force values, resulting in the controller slowly increasing the contact force. To overcome this limitation, relaxation would have to be accounted for in the SFA modelling.

The previously-described linearisation of the SFA pressure-volume relationship is another source of error. Whilst it can be seen that deviations from a linear trend are small for the investigated subset of the inflation range, the mean error associated is 0.06N. Additionally, the assumption of constant force transmissibility for said workspace subset induces errors of approximately 6.69\%.

\section{Conclusion}

This work derived approaches to intrinsic force sensing and control for fluid-driven soft robots. This is achieved incrementally from static sensing, which had first been shown in our previous work. An investigation into the mechanical characteristics of SFAs is provided and limitations in force sensing are identified. The intrinsic force sensing capabilities of the system have been investigated and corresponding mechanical characteristics such as the system's force transmissibility are verified. It has been found that for a single actuator the external force can be sensed with an accuracy of 0.56$\pm$0.66N in a quasi-static configuration. The intrinsic force sensing approach has been applied in a closed-loop force controller for a single SFA. For static contacts the force can be controlled with a steady state accuracy of 0.47$\pm$0.24N. The controller has been expanded to the SEE which is comprised of three SFAs arranged in a parallel, angled configuration. With this setup it has been shown that forces can be controlled along all three Cartesian axes.



\ifCLASSOPTIONcaptionsoff
  \newpage
\fi



\bibliographystyle{ieeetr}
\bibliography{references.bib}

\begin{thebibliography}{10}

\bibitem{Hughes2016}
J.~Hughes, U.~Culha, F.~Giardina, F.~Guenther, A.~Rosendo, and F.~Iida, ``{Soft
  manipulators and grippers: A review},'' 11 2016.

\bibitem{Runciman2019}
M.~Runciman, A.~Darzi, and G.~P. Mylonas, ``{Soft Robotics in Minimally
  Invasive Surgery},'' {\em Soft Robotics}, vol.~6, no.~4, pp.~423--443, 2019.

\bibitem{Back2018}
J.~Back, L.~Lindenroth, K.~Rhode, and H.~Liu, ``{Three dimensional force
  estimation for steerable catheters through bi-point tracking},'' {\em Sensors
  and Actuators, A: Physical}, 2018.

\bibitem{Galloway2016}
K.~C. Galloway, K.~P. Becker, B.~Phillips, J.~Kirby, S.~Licht, D.~Tchernov,
  R.~J. Wood, and D.~F. Gruber, ``{Soft Robotic Grippers for Biological
  Sampling on Deep Reefs},'' {\em Soft Robotics}, vol.~3, no.~1, pp.~23--33,
  2016.

\bibitem{Bishop-Moser2019}
J.~Bishop-Moser, ``{High force generation using inflatable toroidal soft robot
  actuators},'' {\em RoboSoft 2019 - 2019 IEEE International Conference on Soft
  Robotics}, pp.~222--226, 2019.

\bibitem{Yap2016}
H.~K. Yap, H.~Y. Ng, and C.~H. Yeow, ``{High-Force Soft Printable Pneumatics
  for Soft Robotic Applications},'' {\em Soft Robotics}, vol.~3, no.~3,
  pp.~144--158, 2016.

\bibitem{Kosawa2019}
H.~Kosawa and S.~Konishi, ``{Reinforcement Design for Newton-Level High Force
  Generated by Bending Motion of Soft Microactuator},'' {\em 2019 20th
  International Conference on Solid-State Sensors, Actuators and Microsystems
  and Eurosensors XXXIII, TRANSDUCERS 2019 and EUROSENSORS XXXIII}, no.~June,
  pp.~2531--2534, 2019.

\bibitem{Zhang2020}
J.~Zhang, H.~Wei, Y.~Shan, P.~Li, Y.~Zhao, L.~Qi, and H.~Yu, ``{Modeling and
  Experimental Study of a Novel Multi-DOF Parallel Soft Robot},'' {\em IEEE
  Access}, vol.~8, pp.~62932--62942, 2020.

\bibitem{Lindenroth2017}
L.~Lindenroth, A.~Soor, J.~Hutchinson, A.~Shafi, J.~Back, K.~Rhode, and H.~Liu,
  ``{Design of a soft, parallel end-effector applied to robot-guided ultrasound
  interventions},'' in {\em IEEE International Conference on Intelligent Robots
  and Systems}, vol.~2017-Septe, pp.~3716--3721, 2017.

\bibitem{LindenrothTBME}
L.~Lindenroth, R.~J. Housden, S.~Wang, J.~Back, K.~Rhode, and H.~Liu, ``{Design
  and integration of a parallel, soft robotic end-effector for extracorporeal
  ultrasound},'' {\em IEEE Transactions on Biomedical Engineering}, pp.~1--1,
  12 2019.

\bibitem{Shiva2016}
A.~Shiva, A.~Stilli, Y.~Noh, A.~Faragasso, I.~D. Falco, G.~Gerboni,
  M.~Cianchetti, A.~Menciassi, K.~Althoefer, and H.~A. Wurdemann,
  ``{Tendon-Based Stiffening for a Pneumatically Actuated Soft Manipulator},''
  {\em IEEE Robotics and Automation Letters}, vol.~1, pp.~632--637, 7 2016.

\bibitem{Sun2020}
C.~Sun, L.~Chen, J.~Liu, J.~S. Dai, and R.~Kang, ``{A hybrid continuum robot
  based on pneumatic muscles with embedded elastic rods},'' {\em Proceedings of
  the Institution of Mechanical Engineers, Part C: Journal of Mechanical
  Engineering Science}, vol.~234, no.~1, pp.~318--328, 2020.

\bibitem{Tang2020}
Y.~Tang, Y.~Chi, J.~Sun, T.-H. Huang, O.~H. Maghsoudi, A.~Spence, J.~Zhao,
  H.~Su, and J.~Yin, ``{Leveraging elastic instabilities for amplified
  performance: Spine-inspired high-speed and high-force soft robots},'' {\em
  Science Advances}, vol.~6, no.~19, p.~eaaz6912, 2020.

\bibitem{Puangmali2008}
P.~Puangmali, K.~Althoefer, L.~D. Seneviratne, D.~Murphy, and P.~Dasgupta,
  ``{State-of-the-Art in Force and Tactile Sensing for Minimally Invasive
  Surgery},'' {\em IEEE Sensors Journal}, vol.~8, pp.~371--381, 4 2008.

\bibitem{Hoffmayer2015}
K.~S. Hoffmayer and E.~P. Gerstenfeld, ``{Contact force-sensing catheters},''
  {\em Current Opinion in Cardiology}, vol.~30, no.~1, pp.~74--80, 2015.

\bibitem{Polygerinos2009}
P.~Polygerinos, T.~Schaeffter, L.~Seneviratne, and K.~Althoefer, ``{Measuring
  tip and side forces of a novel catheter prototype: A feasibility study},''
  {\em 2009 IEEE/RSJ International Conference on Intelligent Robots and
  Systems, IROS 2009}, pp.~966--971, 2009.

\bibitem{Kesner2011}
S.~B. Kesner and R.~D. Howe, ``{Force control of flexible catheter robots for
  beating heart surgery},'' {\em Proceedings - IEEE International Conference on
  Robotics and Automation}, pp.~1589--1594, 2011.

\bibitem{Xiao2012}
N.~Xiao, J.~Guo, S.~Guo, and T.~Tamiya, ``{A robotic catheter system with
  real-time force feedback and monitor},'' {\em Australasian Physical and
  Engineering Sciences in Medicine}, vol.~35, no.~3, pp.~283--289, 2012.

\bibitem{Polygerinos2013}
P.~Polygerinos, L.~D. Seneviratne, R.~Razavi, T.~Schaeffter, and K.~Althoefer,
  ``{Triaxial catheter-tip force sensor for MRI-guided cardiac procedures},''
  {\em IEEE/ASME Transactions on Mechatronics}, vol.~18, no.~1, pp.~386--396,
  2013.

\bibitem{Paridah2016}
M.~Paridah, A.~Moradbak, A.~Mohamed, F.~a.~t. Owolabi, M.~Asniza, and S.~H.
  Abdul~Khalid, ``{Microsphere and Fiber Optics based Optical Sensors},'' in
  {\em Intech}, vol.~i, p.~13, 2016.

\bibitem{Xie2015}
H.~Xie, H.~Liu, Y.~Noh, J.~Li, S.~Wang, and K.~Althoefer, ``{A
  fiber-optics-based body contact sensor for a flexible manipulator},'' {\em
  IEEE Sensors Journal}, vol.~15, no.~6, pp.~3543--3550, 2015.

\bibitem{Shiva2019}
A.~Shiva, S.~M. Sadati, Y.~Noh, J.~Fra{\'{s}}, A.~Ataka, H.~W{\"{u}}rdemann,
  H.~Hauser, I.~D. Walker, T.~Nanayakkara, and K.~Althoefer, ``{Elasticity
  Versus Hyperelasticity Considerations in Quasistatic Modeling of a Soft
  Finger-Like Robotic Appendage for Real-Time Position and Force Estimation},''
  {\em Soft Robotics}, vol.~6, no.~2, pp.~228--249, 2019.

\bibitem{Qasaimeh2009}
M.~A. Qasaimeh, S.~Sokhanvar, J.~Dargahi, and M.~Kahrizi, ``{PVDF-based
  microfabricated tactile sensor for minimally invasive surgery},'' {\em
  Journal of Microelectromechanical Systems}, vol.~18, no.~1, pp.~195--207,
  2009.

\bibitem{Seminara2013PiezoelectricSensors}
L.~Seminara, L.~Pinna, M.~Valle, L.~Basirico, A.~Loi, P.~Cosseddu,
  A.~Bonfiglio, A.~Ascia, M.~Biso, A.~Ansaldo, D.~Ricci, and G.~Metta,
  ``{Piezoelectric polymer transducer arrays for flexible tactile sensors},''
  {\em IEEE Sensors Journal}, vol.~13, no.~10, pp.~4022--4029, 2013.

\bibitem{Chossat2013}
J.~B. Chossat, Y.~L. Park, R.~J. Wood, and V.~Duchaine, ``{A soft strain sensor
  based on ionic and metal liquids},'' {\em IEEE Sensors Journal}, vol.~13,
  no.~9, pp.~3405--3414, 2013.

\bibitem{Dickey2017}
M.~D. Dickey, ``{Stretchable and Soft Electronics using Liquid Metals},'' {\em
  Advanced Materials}, vol.~29, no.~27, pp.~1--19, 2017.

\bibitem{Cooper2017}
C.~B. Cooper, K.~Arutselvan, Y.~Liu, D.~Armstrong, Y.~Lin, M.~R. Khan,
  J.~Genzer, and M.~D. Dickey, ``{Stretchable Capacitive Sensors of Torsion,
  Strain, and Touch Using Double Helix Liquid Metal Fibers},'' {\em Advanced
  Functional Materials}, vol.~27, no.~20, 2017.

\bibitem{Kramer2015}
R.~K. Kramer, ``{Soft electronics for soft robotics},'' {\em Proc. SPIE 9467,
  Micro- and Nanotechnology Sensors, Systems, and Applications VII}, vol.~9467,
  p.~946707, 2015.

\bibitem{Khoshnam2015}
M.~Khoshnam, A.~C. Skanes, and R.~V. Patel, ``{Modeling and estimation of tip
  contact force for steerable ablation catheters},'' {\em IEEE Transactions on
  Biomedical Engineering}, vol.~62, no.~5, pp.~1404--1415, 2015.

\bibitem{Rucker2011}
D.~C. Rucker and R.~J. Webster, ``{Deflection-based force sensing for continuum
  robots: A probabilistic approach},'' {\em IEEE International Conference on
  Intelligent Robots and Systems}, pp.~3764--3769, 2011.

\bibitem{Yuan2017a}
H.~Yuan, P.~W.~Y. Chiu, and Z.~Li, ``{Shape-Reconstruction-Based Force Sensing
  Method for Continuum Surgical Robots with Large Deformation},'' {\em IEEE
  Robotics and Automation Letters}, 2017.

\bibitem{Zhang2018}
Z.~Zhang, J.~Dequidt, and C.~Duriez, ``{Vision-Based Sensing of External Forces
  Acting on Soft Robots Using Finite Element Method},'' {\em IEEE Robotics and
  Automation Letters}, vol.~3, no.~3, pp.~1529--1536, 2018.

\bibitem{Back2016}
J.~Back, L.~Lindenroth, R.~Karim, K.~Althoefer, K.~Rhode, and H.~Liu, ``{New
  kinematic multi-section model for catheter contact force estimation and
  steering},'' in {\em IEEE International Conference on Intelligent Robots and
  Systems}, vol.~2016-Novem, pp.~2122--2127, 2016.

\bibitem{Xu2008}
K.~Xu and N.~Simaan, ``{An investigation of the intrinsic force sensing
  capabilities of continuum robots},'' {\em IEEE Transactions on Robotics},
  vol.~24, no.~3, pp.~576--587, 2008.

\bibitem{Xu2010a}
K.~Xu and N.~Simaan, ``{Intrinsic wrench estimation and its performance index
  for multisegment continuum robots},'' {\em IEEE Transactions on Robotics},
  vol.~26, no.~3, pp.~555--561, 2010.

\bibitem{Haraguchi2011}
D.~Haraguchi, K.~Tadano, and K.~Kawashima, ``{A prototype of
  pneumatically-driven forceps manipulator with force sensing capability using
  a simple flexible joint},'' {\em IEEE International Conference on Intelligent
  Robots and Systems}, pp.~931--936, 2011.

\bibitem{Black2018}
C.~B. Black, J.~Till, and D.~C. Rucker, ``{Parallel Continuum Robots: Modeling,
  Analysis, and Actuation-Based Force Sensing},'' {\em IEEE Transactions on
  Robotics}, vol.~34, no.~1, pp.~29--47, 2018.

\bibitem{Lindenroth2017b}
L.~Lindenroth, C.~Duriez, J.~Back, K.~Rhode, and H.~Liu, ``{Intrinsic force
  sensing capabilities in compliant robots comprising hydraulic actuation},''
  in {\em IEEE International Conference on Intelligent Robots and Systems},
  2017.

\end{thebibliography}
\end{document}